\documentclass[10pt,twocolumn,letterpaper]{article}

\usepackage[pagenumbers]{iccv} 

\usepackage[ruled,vlined, linesnumbered]{algorithm2e}
\usepackage{graphicx}
\usepackage{algorithmic}
\usepackage{pifont}
\usepackage{subcaption}
\usepackage{multirow}

\usepackage{tikz}
\usepackage{graphicx}

\newcommand{\cmark}{\ding{51}}%
\newcommand{\xmark}{\ding{55}}%

\definecolor{iccvblue}{rgb}{0.21,0.49,0.74}
\usepackage[pagebackref,breaklinks,colorlinks,allcolors=iccvblue]{hyperref}

\def\paperID{5009} 
\def\confName{ICCV}
\def\confYear{2025}

\title{DDB: Diffusion Driven Balancing to Address Spurious Correlations}

\author{
Aryan Yazdan Parast, Basim Azam, Naveed Akhtar \\
The University of Melbourne, Australia \\
 \\
{\tt\small ayazdanparas@student.unimelb.edu.au,  \{basim.azam, naveed.akhtar1\}@unimelb.edu.au}
}

\begin{document}
\maketitle

\begin{abstract}
Deep neural networks trained with Empirical Risk Minimization (ERM) perform well when both training and test data come from the same domain, but they often fail to generalize to out-of-distribution samples. In image classification, these models may rely on spurious correlations that often exist between labels and irrelevant features of images, making predictions unreliable when those features do not exist. We propose a Diffusion Driven Balancing (DDB) technique to generate training samples with text-to-image diffusion models for addressing the spurious correlation problem. First, we compute the best describing token for the visual features pertaining to the causal components of samples by a textual inversion mechanism. Then, leveraging a language segmentation method and a diffusion model, we generate new samples by combining the causal component with the elements from other classes. We also meticulously prune the generated samples based on the prediction probabilities and attribution scores of the ERM model to ensure their correct composition for our objective. Finally, we retrain the ERM model on our augmented dataset. This process reduces the model’s reliance on spurious correlations by learning from carefully crafted samples in which this correlation does not exist. Our experiments show that across different benchmarks, our technique achieves better worst-group accuracy than the existing state-of-the-art methods. Our code is available at \url{https://github.com/ArianYp/DDB}.
\end{abstract}
\vspace{-3mm}
\section{Introduction}
\vspace{-1mm}
Deep neural networks trained with Empirical Risk Minimization (ERM) do not perform well when the train and test datasets do not follow the same distribution \cite{arjovsky2020invariantriskminimization, hovy-sogaard-2015-tagging, DBLP:journals/corr/abs-1711-11561}. They may rely on data features that correlate to the sample labels but are not stable across different domains, leading to poor model generalization \cite{arjovsky2020invariantriskminimization, DBLP:journals/corr/abs-1711-11561, noohdani2024decomposeandcomposecompositionalapproachmitigating}. For example, a model  trained to classify camels and cows may learn to rely on background cues rather than the animals themselves. This occurs because most cows appear in green pastures, while most camels are found in deserts. Hence, the model may learn to classify images based solely on the background, and fail when this spurious correlation is absent, e.g., when it encounters a camel in a green pasture or a cow in a desert~\cite{arjovsky2020invariantriskminimization, beery2018recognitionterraincognita}. 

The problem is commonly handled by assuming that data samples belong to different subgroups, each representing a distinct domain. The solutions then seek to maximize the model accuracy for the worst group under this perspective~\cite{sagawa2020distributionallyrobustneuralnetworks, liu2021justtraintwiceimproving}.
To that end, multiple approaches aim at computing robust representations that are stable across different domains \cite{arjovsky2020invariantriskminimization, ahmed2021systematic}. Other methods  balance between the groups in the dataset by reweighting \cite{liu2021justtraintwiceimproving, qiu2023simplefastgrouprobustness}, combining or mixing  images \cite{wu2023discovercureconceptawaremitigation, noohdani2024decomposeandcomposecompositionalapproachmitigating, yao2022improvingoutofdistributionrobustnessselective} or generating new samples \cite{yuan2024justprettypicturesinterventional, qraitem2024fakerealpretrainingbalanced}. Reweighting methods \cite{liu2021justtraintwiceimproving, qiu2023simplefastgrouprobustness, nam2020learningfailuretrainingdebiased} assume that minority groups have high loss, hence they upweight those groups. However, in ERM models, a high loss can arise from  reasons other than the group imbalance, leading to poor identification of the minority group. Additionally, the number of minority group samples can be too low to effectively mitigate the problem with reweighting.

To improve the minority groups,  methods such as \cite{wu2023discovercureconceptawaremitigation, noohdani2024decomposeandcomposecompositionalapproachmitigating, yao2022improvingoutofdistributionrobustnessselective} attempt to augment data manually by mixing \cite{wu2023discovercureconceptawaremitigation, noohdani2024decomposeandcomposecompositionalapproachmitigating} or interpolating \cite{yao2022improvingoutofdistributionrobustnessselective} between the samples. Noohdani et al.~\cite{noohdani2024decomposeandcomposecompositionalapproachmitigating} viewed the problem through a compositional lens, where  images are assumed to be  composed of causal and non-causal parts. Under this view,  existence of (spurious) correlation between a sample's non-causal part and its label causes its misclassification. 
DaC \cite{noohdani2024decomposeandcomposecompositionalapproachmitigating} aims to identify both causal and non-causal parts of images and leverages them to augment  the minority groups. However, lack of semantic control in the augmentation samples and their poor composition quality still hinder the performance of this method.

\begin{figure*}
    \centering
    \includegraphics[width=\linewidth]{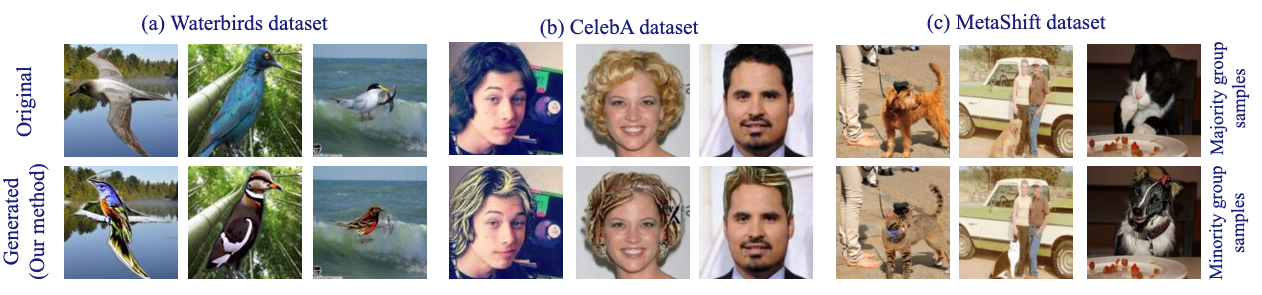}
    \vspace{-7mm}
    \caption{Examples of generated images by the proposed approach. We perform high-quality automated modification to the majority group samples (top) to generate minority group samples (bottom) such that the alterations precisely occur to the causal features in a manner that resolves the spurious correlations.     
    (a) For Waterbirds dataset~\cite{article, sagawa2020distributionallyrobustneuralnetworks}, which contains majority samples of landbirds with land backgrounds and waterbirds with water backgrounds, our method generates new images with landbirds on water backgrounds and waterbirds on land backgrounds while retaining majority sample backgrounds. (b) For  CelebA~\cite{liu2015deeplearningfaceattributes}, where the majority consists of non-blond males and blond females, our approach generates images of blond males and non-blond females. (c) For the MetaShift dataset~\cite{liang2022metashiftdatasetdatasetsevaluating}, where dogs and cats are correlated with specific objects in the background, our technique  generates samples that break this correlation by swapping dogs and cats.} 
    \label{fig:grouped_datasets_adjusted}
    \vspace{-3mm}
\end{figure*}

Recently, diffusion-based generative models~\cite{ho2020denoisingdiffusionprobabilisticmodels} such as Stable Diffusion (SD) \cite{rombach2022highresolutionimagesynthesislatent},
 have demonstrated  high quality image generation using prompts. Studies affirm that their generative images help in  few- and zero-shot recognition~\cite{he2023syntheticdatagenerativemodels}. Consequently, there is also an  interest in exploiting them to enhance models for out-of-distribution generalization~\cite{yuan2024justprettypicturesinterventional, qraitem2024fakerealpretrainingbalanced, dunlap2023diversifyvisiondatasetsautomatic}. 
Along similar lines, we aim at addressing the spurious correlation problem through group balancing by leveraging  generative models~\cite{ho2020denoisingdiffusionprobabilisticmodels}. We devise a unique guidance strategy for our generative process to enable sampling from the minority group distribution with a data-centric approach that is largely agnostic to prompts, and human intervention by extension. 


We refer to our method as Diffusion Driven Balancing (DDB).
Inline with  the compositional view of images~\cite{noohdani2024decomposeandcomposecompositionalapproachmitigating}, it leverages  
textual inversion \cite{gal2022imageworthwordpersonalizing} and identifies the optimal tokens for the causal parts of each class.  It then employs language-based segmentation  \cite{kirillov2023segment} to identify the visual causal counterparts in the images. These parts are later employed by our technique for diffusion-based image generation that can balance the data. In  Fig.~\ref{fig:grouped_datasets_adjusted}, we show representative examples of the  generated  minority group samples computed by our technique by altering the causal parts of the majority class images from the training data.
A key challenge in generating such samples  is ensuring that they truly capture the desired properties for addressing the spurious correlation problem. Otherwise, the impact of balancing can be detrimental. Our processing pipeline leverages multiple components, and under-performance by any of those can potentially lead to a low quality image. 
To handle that, we introduce an effective pruning strategy based on the attribution score and the prediction probability distribution of the ERM model. Our approach evaluates how much a generated sample influences the model’s predictions, ensuring that only samples with a strong causal impact are retained. Moreover, by further leveraging attribution scores, our approach ensures higher-quality relevant images, leading to improved group balancing and better generalization.

Our contributions are summarized as follows.
\begin{itemize}
\item We propose a unique automated diffusion-based technique to generate samples for minority groups to address spurious correlation problem for image classifiers. Exploiting compositional nature of images and class label information, our method precisely modifies the available samples for improved ERM model performance.
\item Our method adapts token inversion, language-based segmentation and diffusion-based image inpainting, while also introducing a unique data pruning strategy to ensure retention of high quality augmentation samples. 
\item We extensively evaluate our method on multiple benchmarks. Our approach outperforms existing methods and demonstrates promises for sophisticated diffusion based frameworks for mitigating spurious correlations.
\end{itemize}

\section{Related work}
\label{sec:related}
Our eventual goal is to achieve group robustness and enable improved worst-group accuracy for the models. Existing methods pursuing the same objective can be categorized into augmention-based and   non-augmentation-based techniques. We discuss both categories below. 

\subsection{Group robustness without data augmentation}
There are existing techniques that aim at achieving group robustness by modifying the training procedure of the model~\cite{sagawa2020distributionallyrobustneuralnetworks, arjovsky2020invariantriskminimization}. The IRM method~\cite{arjovsky2020invariantriskminimization} seeks to learn a representation that is robust to correlation shift by incorporating an invariant risk minimization loss term in training. G-DRO~\cite{sagawa2020distributionallyrobustneuralnetworks} minimizes the worst-group loss across different groups, requiring access to group labels during training.  

Another line of approaches focuses on reweighting the minority samples during model training~\cite{liu2021justtraintwiceimproving, nam2020learningfailuretrainingdebiased, qiu2023simplefastgrouprobustness}. For instance,  JTT \cite{liu2021justtraintwiceimproving} increases the weight of misclassified samples from an initial model trained with ERM for a small number of epochs. The LFF method~\cite{nam2020learningfailuretrainingdebiased} employs an auxiliary biased model 
to assist in training a more robust model. 
Specifically, it assigns higher weights to the samples that the former model struggles to learn, encouraging the latter model to focus on those challenging examples. The AFR approach~\cite{qiu2023simplefastgrouprobustness} also assigns weights to each sample in the dataset based on its incorrectness score computed from the ERM model. Inspired by these approaches, we leverage the ERM loss and classification outputs to intuitively identify majority and minority groups. DFR \cite{kirichenko2023layerretrainingsufficientrobustness} showed that retraining only the classifier layer of an ERM model on a balanced dataset is sufficient to achieve group robustness.

\subsection{Data augmentation-based methods}

A line of approaches \cite{noohdani2024decomposeandcomposecompositionalapproachmitigating, qraitem2024fakerealpretrainingbalanced, wu2023discovercureconceptawaremitigation, yao2022improvingoutofdistributionrobustnessselective, taghanaki2022masktunemitigatingspuriouscorrelations} addresses group robustness by balancing groups through data augmentation.  
For instance, MaskTune \cite{taghanaki2022masktunemitigatingspuriouscorrelations} assumes that the attention of an ERM model focuses on spurious parts of an image. Based on this, it masks these regions and retrains the ERM model for one epoch on the masked images. DISC \cite{wu2023discovercureconceptawaremitigation} first identifies spurious attributes in each class using an auxiliary concept bank and then applies mixup to balance spurious attributes across different classes. LISA \cite{yao2022improvingoutofdistributionrobustnessselective} performs data augmentation by interpolating samples and labels. The samples for interpolation come either from the same domain but with different labels or from different domains while sharing the same label.  DaC \cite{noohdani2024decomposeandcomposecompositionalapproachmitigating} generates samples from minority groups by combining causal and non-causal parts of images from different classes. It disentangles these parts based on the attribution score from the ERM model.  The FFR method~\cite{qraitem2024fakerealpretrainingbalanced}  is the most similar work to ours, as it also utilizes diffusion models to address dataset imbalance. It first generates a balanced synthetic dataset and trains the model on that to learn a robust representation for each group. Then, it retrains the model on the real dataset to better fit to its distribution. However, this method is sensitive to prompts, and it can lead to augmentation samples  detrimental to the cause.
Moreover, it requires knowledge about the bias present in the dataset. Other related works, e.g.,  \cite{yuan2024justprettypicturesinterventional, vendrow2023datasetinterfacesdiagnosingmodel, dunlap2023diversifyvisiondatasetsautomatic}, focus on diversifying samples in the dataset by augmenting data with generative models. However, none of them explicitly investigate group robustness or evaluate the technique for worst-group accuracy following the standard spurious correlation benchmarks.  

\vspace{-2mm}
\section{Problem setup}
\label{setup}
\vspace{-1mm}
Given a dataset $D_{tr} = \{(x_i, y_i)\}_{i=0}^{N}$, a typical  classifier $f$ is aimed at estimating $P(y|x)$.
We model the problem with Structrul Causal 
We consider the compositional nature~\cite{yao2022improvingoutofdistributionrobustnessselective} of the samples $\{x_i\}_{i=0}^{N}$, where any sample consists of two types of features: core features ($c$) and spurious features ($s$).
Core features (c) causally determine the correct label of the sample, while the spurious features (s) correlate with the label mainly due to irrelevant peculiarities of  the dataset, e.g., class imbalances. These features are not inherently predictive. In other words, $P(y|s) \neq P(y)$.  

To clarify, we provide a quick example from  CelebA dataset~\cite{liu2015deeplearningfaceattributes}. The objective in CelebA is to classify images of celebrities as blond or non-blond. Hence, the hair color serves as the core feature. However, other attributes, such as gender (male/female) and hair length (short/long), may correlate with the label (blond/non-blond) due to the  peculiar composition of the dataset. Specifically, most images of blond individuals in the dataset are of women with long hair.  This leads to a spurious correlation between female gender and long hair with the blond hair color. 

A dataset can be partitioned into majority and minority groups based on $(c, s)$ combinations, where majority samples contain a spurious feature ($s_{\text{maj}}$). In the Structural Causal Model (SCM)~\cite{Pearl_2009} view, this creates a backdoor path $s_{\text{maj}} \rightarrow Y$, leading the classifier to rely on those features. Consequently, it fails to generalize when the correlation breaks \cite{xiao2020noisesignalroleimage, geirhos2022imagenettrainedcnnsbiasedtexture}. Following common convention, we refer to the group on which $f$'s performance is worst as the \textit{worst group}.


\vspace{-2mm}
\section{Method}
\begin{figure*}[t]
    \centering
    \includegraphics[width = \linewidth]{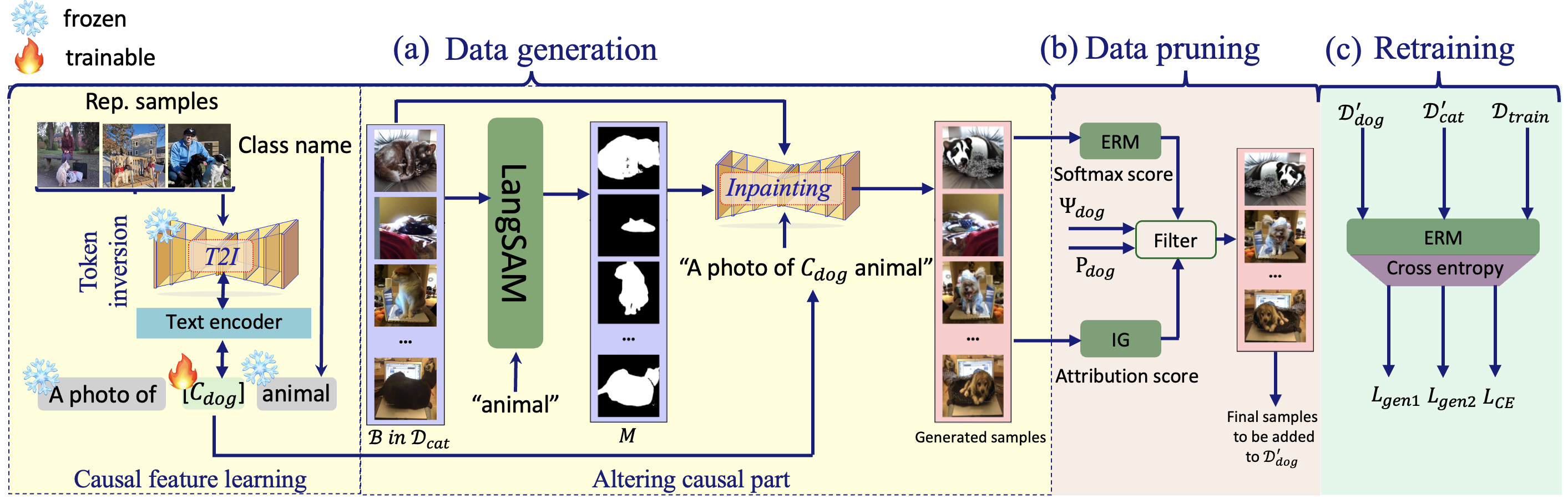}
    \vspace{-5mm}
    \caption{An overview of DDB - Dog/Cat classes used for illustration. (a) First, the token $C_{dog}$ is optimized by reconstructing samples from the \textit{Dog} class using prompts that include the trainable $C_{dog}$. Then, for each batch in $\mathcal{D}_{cat}$, masks $M$ are generated using the LangSAM~\cite{kirillov2023segment} with the prompt \textit{Class Name = `animal'}. These masks, along with the input images and a textual prompt, are fed into the inpainting model to generate new images. The token $C_{dog}$ is incorporated to generate the causal features of the \textit{Dog} class. (b) The generated samples are then passed through the ERM model and the integrated gradients (IG) module to compute relevant scores for filtering undesired generated images.  Algorithm for this stage is detailed in Alg.~\ref{alg:psudocode_generate}. (c) The final set of generated samples is used to retrain the ERM model.}
    \label{fig:pipeline}
    \vspace{-3mm}
\end{figure*}

\vspace{-1mm}
We introduce DDB (Diffusion-Driven-Balancing), a three-stage approach designed to balance dataset groups effectively - see Fig.~\ref{fig:pipeline}. Our first stage focuses on generating new data for minority groups using a diffusion-based  generative model~\cite{rombach2022highresolutionimagesynthesislatent}. In the second stage, the generated data is pruned based on empirical risk minimization (ERM) loss and attribution scores, ensuring that only high-quality and relevant samples are retained. Finally, in the third stage, the ERM model is retrained on the newly balanced dataset, improving its robustness and mitigating spurious correlations. Below, we discuss these stages in detail.

\vspace{-1mm}
\subsection{Generating data}
\label{3.1}

\textbf{Causal feature learning}. 
Trained on enormous datasets, contemporary text-to-image (T2I) generative models can provide strong priors for natural image data augmentation~\cite{croitoru2023diffusion}.   
We leverage this fact  to synthesize underrepresented samples in training data for addressing the spurious correlation problem. 
To that end, we choose Stable Diffusion (SD) \cite{ho2020denoisingdiffusionprobabilisticmodels} for its established performance and accessibility.
Our objective of addressing spurious correlation with SD necessitates a precise control over the (core/spurious) features of the generated content, which is not readily available in the original SD. 
Hence, we propose to first  learn special token embeddings by tuning a text encoder to gain the ability of generating causal features  for each class. Inspired by textual inversion~\cite{gal2022imageworthwordpersonalizing}, we generate label-$i$ samples using templates that include a learnable token $[C_i]$, with the class name placed at the end to emphasize the image's causal semantics. For instance, in the Waterbirds dataset - a popular benchmark for the problem (see \S~\ref{sec:settings}), where the class name is \texttt{"bird"},  our template sentence for the LandBird label would be: 
\begin{center}
   $ \texttt{"A photo of a } [C_{\text{landbird}}] \texttt{ bird."}$
\end{center}
where $[C_{\text{landbird}}]$ is a learnable token representing the causal features of the Landbird category. 

Concretely, our method employs publicly available SD version-2~\cite{rombach2022highresolutionimagesynthesislatent} as the T2I model while keeping its parameters frozen. To gain  control over the causal features, we select a random subset of the samples from each class and  tune the text encoder by minimizing the following loss, which leads to learning an optimized embedding of the token. 
\begin{equation}
    C^*_{i} \!=\! \arg\min_{C_{i}} \mathbb{E}_{z \sim \mathcal{E}(x), I, \epsilon \sim \mathcal{N}(0,1), t} 
    \!\left[ \|\epsilon - \epsilon_{\theta}(z_t, t, \tau_{\theta}(I))\|_2^2 \right],
    \label{eq:TokenInversion}
\end{equation}
where $\tau_{\theta}(I)$ is the text encoder model, $I$ is the input prompt, $z_t$ is the latent noisy representation at time step $t$, $\epsilon$ is a noise sample,  $\epsilon_\theta$ is the denoising network, and input image $x$ from class $i$. It is also noted that SD operates in the latent space of an auto-encoder. In Eq.~(\ref{eq:TokenInversion}), $\mathcal{E(.)}$ denotes the encoder of that model, which remains frozen in our technique.   The learned token for the causal feature eventually gets incorporated into the input template sentences of our generative process to serve as a conditioning signal for SD -  see Fig.~\ref{fig:pipeline}~(a), which we use to alter causal parts of images. 

\vspace{0.5mm}
\noindent\textbf{Altering causal part}. From the compositional perspective, we aim to generate underrepresented samples in the data by selectively modifying the constituting components of images. 
Under this perspective, the minority samples contain  spurious parts of the majority group from another class.  To upsample the minority groups, we must modify the causal parts of the majority group samples, transforming them into another class while preserving the spurious background - see  Fig.~\ref{fig:grouped_datasets_adjusted}(bottom).
To achieve this, we selectively add noise only to the causal parts of images and denoise them using the tokens $C^*$ learned in the previous stage. 
We utilize the inpainting abilities of SD using a masking function
$m$ to restrict the noising area. 

Given a masking function \( m \), let \( M = m(x_j) \) be a binary mask for an image \( x_j \), identifying its causal part. Our inpainting  model then applies noise only to the masked region of the image. The forward pass of the underlying diffusion process can be expressed as 
\begin{equation}
    z_t = (1 - M) \odot z_0 + M \odot \left( \sqrt{\bar{\alpha}_t} z_0 + \sqrt{1 - \bar{\alpha}_t} \epsilon \right),
    \label{eq:maskedDDPM}
\end{equation}
where \( z_t \) is the latent noisy representation at time step \( t \), \( z_0 \) is the original latent representation, and \( \epsilon \sim \mathcal{N}(0,1) \) is the noise sample. In Eq.~(\ref{eq:maskedDDPM}), the application of the mask along the additional term of $(1-M) \odot z_0$ determines  the forward diffusion trajectory only based on   the masked region. Hence,  the denoising process can be focused on generating the  masked region while the unmasked region remains unchanged for the output image. 

We exploit the language-driven image segmentation abilities of  LangSAM \cite{kirillov2023segment}\footnote{Language Segment Anything Model (\url{https://github.com/luca-medeiros/lang-segment-anything}) is a language-driven image segmentation model that combines the strengths of GroundingDINO \cite{liu2024groundingdinomarryingdino} and Segment Anything Model (SAM)  \cite{kirillov2023segment}.}
to compute the mask $M$ in our approach. 
This method enables a convenient automation of the masking function $m(x_j): x_j \rightarrow M$ discussed above. 
As input, LangSAM needs an image and a text prompt for the object to be segmented. The prompt is readily available for our problem in the form of [Class-name] for any available image subset $\mathcal{D}_{\text{class-name}}$.
Hence, LangSAM gets  easily  adopted in our pipeline shown in Fig.~\ref{fig:pipeline}. 

\begin{algorithm}
  \caption{Generate New Dataset}
  \label{alg:psudocode_generate}
  \SetAlgoLined
  \KwIn{Model $f_{\theta}$; Class $i$, $i'$, Mean softmax $\Psi_i$; Threshold $\mathrm{P}_{i'}$; Masking Function $m$, Token $C^*_{i'}$, Dataset $\mathcal{D}_{i}$, Inpainting model $SD$ }
  \KwOut{Dataset $\mathcal{D}'_{i'}$}
  $\mathcal{D}'_{i'} \leftarrow \{\}$ \;
  $prompt \leftarrow \texttt{A photo of a \( C^*_{i'} \) [Class name]}$ \;
  \For{batch $\mathcal{B}$ in $\mathcal{D}_{i}$}{
    $M \leftarrow m(\mathcal{B)}$ \;
    $generated\_samples \leftarrow SD(\mathcal{B}, M, prompt)$\;
    \For{$x' \in generated\_samples, x\in\mathcal{B}$} 
    {  $\psi \leftarrow \text{Softmax}(f_\theta, x',y)$\;
  \tcp{Equation (\ref{eq6})} 
  $IG_{score} \leftarrow \text{IG}(f_\theta, x',x,y')$\; 
  $\rho \leftarrow \sum_{k} M_k IG_{score_k}$ \;
    \If{$\psi < \Psi_i$ \textbf{and} $\rho > \mathrm{P}_{i'}$}{    
      $\mathcal{D'}_{i'}\leftarrow \mathcal{D'}_{i'} \cup \{(x',y')\}$
  }
    }
  }
  \Return $\mathcal{D'}_{i'}$
\end{algorithm}

Since we are after altering the causal parts of the majority group, identification of that group is also required by our technique. Instead of relying on manual labeling, e.g. in \cite{yao2022improvingoutofdistributionrobustnessselective}, we make use of the observation
that low-loss samples for the available model are  likely to be the majority group samples~\cite{noohdani2024decomposeandcomposecompositionalapproachmitigating}. Based on this observation, given a dataset \( D_{tr} = \{(x_j, y_j)\}_{j=0}^{N} \) and the ERM model \( f \), we select a subset \( D_{i} \) containing \( K \) samples that have the lowest loss on \( f \) in class \( i \). This selection ensures that almost all samples in \( D_{i} \) are from majority groups, allowing our method to be applied to generate underrepresented samples. For a data point \( (x_j, y_j) \in D_i \), our method obtains the mask \( M_j \) from the LangSAM model and generates a new sample using the prompt \texttt{A photo of a \( C^*_{i'} \) [Class name]}, with \( x_j \) and \( M_j \) as input. The resulting image will be an underrepresented sample from class $i'$.

\begin{figure}[t]
    \centering
    \includegraphics[width=\linewidth]{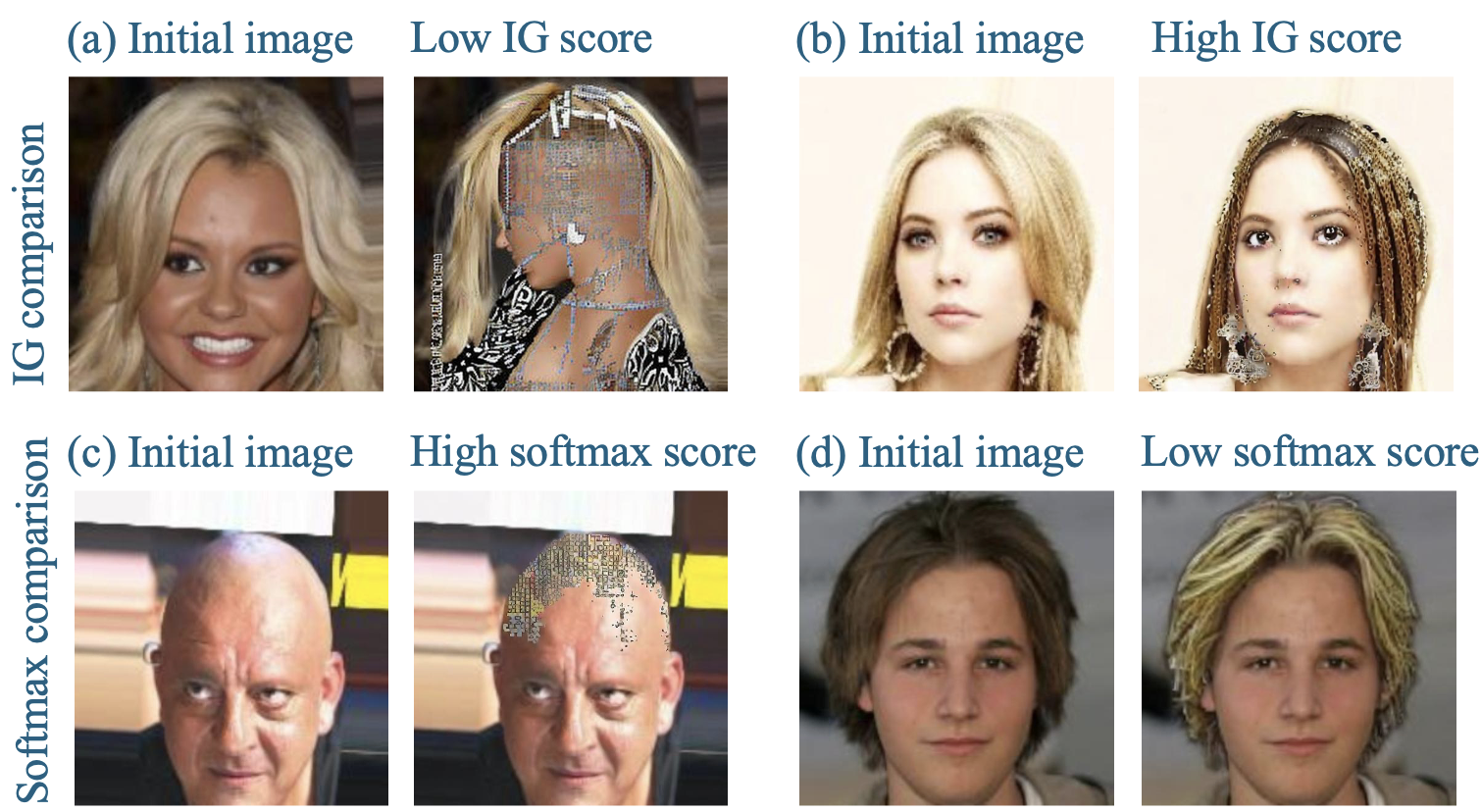}
    \vspace{-5mm}
    \caption{Examples of poor and acceptable scores of the generated images. (a) The diffusion model fails to generate a proper image, and the altered part does not contribute to the label change, resulting in a low IG score. However, the ERM model's prediction changes significantly, requiring pruning. (b) The generated sample has a high IG score because the newly generated features effectively contribute to the label change in the ERM model, no pruning needed. (c) The image remains largely unchanged, leading to no significant shift in the ERM softmax score. However, it has a high IG score, pruning needed. (d) The ERM model prediction changes due to an appropriate level of modification in the image, not pruned.}
    \vspace{-3mm}
    \label{fig:failed_samples}
\end{figure}
\vspace{-2mm}
\subsection{Pruning}
\vspace{-1mm}
\label{3.2}
Despite our meticulous generative pipeline  for balancing the underrepresented samples, the generated samples may contain undesired instances. Hence,  we propose to further prune these samples.
For that, we introduce a bicephalous  pruning condition for  filtering the generated images, leveraging  ERM model's softmax layer and  attribution score.

The ERM model can be denoted as 
$f = w \circ g_\theta$, where $g_\theta$ represents the backbone  and 
$w$ is a linear predictor mapping the output of $g_\theta$ to the number of classes
by applying softmax to obtain a valid probability distribution over the classes.  
In our method, if the diffusion model fails to generate the appropriate visual features or LangSAM is unable to detect the causal part, the ERM model will continue to rely on the spurious cues for its prediction. This is true  because no meaningful features get further  introduced into the image to shift the ERM model’s attention - see Fig.~\ref{fig:failed_samples}(c).
Consequently, its output probabilities will only alter slightly, if at all. Based on this intuition, we exploit the probability assigned to each class as a pruning criterion.

Using the  training samples, we define $\mathcal{D}_{\text{train}_i}$ as the set of all training samples from class $i$. We then compute:
\begin{equation}
    \label{softmax}
    \Psi_i = \frac{1}{|\mathcal{D}_{\text{train}_i}|} \sum_{(x,y) \in \mathcal{D}_{\text{train}_i}} \text{Softmax}(f_{\theta}(x), y)[i].
\end{equation}
After modifying the causal part of a sample \( x_j \) from class \( i \) to generate \( x'_j \), we also compute \(\psi_j =  \text{Softmax}(f_{\theta}(x'_j), y)[i] \).
If $\psi_j \geq \Psi_i$, then the image has not changed desirably and likely still belongs to the majority group of class~\( i \), hence we filter out such images. 

Other potential failure cases for our generative process may result from unintended part inclusion in segmentation by LangSAM or Stable Diffusion generating incorrect causal attributes - see Fig.~\ref{fig:failed_samples}(a). To address these case, we filter images based on the attention of the ERM model to the masked region. 

In our process, we are aiming at altering the causal part of each image such that its predicted label also changes. Consequently, the attribution score of the ERM model should increase for the new label. To compute the attribution scores, we utilize integrated gradients \cite{sundararajan2017axiomaticattributiondeepnetworks}, as it provides precise attributions and quantifies the difference in the ERM model's attention across two images. Integrated gradients compute attributions by following a straight path from a baseline to the image, accumulating gradients at all points along this trajectory.
Using this, we calculate the gradients for both the initial and modified samples, allowing us to assess the impact of the altered image region on the ERM model. This helps determine how much the new image contributes to the new label compared to the original image.

Given a modified sample \( x' \) in the new class \( i' \), an initial sample \( x \) in class \( i \), and an ERM model \( f \) with a prediction score \( f_{i'} \) for class \( i' \), the integrated gradient for the \( k \)th dimension in our method gets  defined as:
\begin{align}
    \label{eq6}
    \text{IG}_k(x') = 
    (x_k' - x_k) \times \int_{0}^{1} \frac{\partial f_{i'} (x + \alpha \times (x' - x))}{\partial x'_k}, d\alpha
\end{align}
where $\alpha$ sweeps the linear path between the images. 
We accumulate all dimensions in the masked region to get an attribution  score for the new sample. Given the mask \( M = m(x) \), we calculate the final attribution score as follows:
\begin{equation}
    \label{Igmean}
    \text{$\rho$} = \sum_{k} M_k \odot IG_k.
\end{equation}
We threshold the score with a hyperparameter \( \mathrm{P} \) to filter out the undesired images.  The overall image generation process with filtration is summarized in Alg \ref{alg:psudocode_generate}.

\vspace{-2mm}
\subsection{Retraining}
\vspace{-1mm}
Given a dataset \( \mathcal{D}_{\text{tr}} = \{(x_j, y_j)\}_{j=0}^{N} \) and the ERM model \( f \), we generate new   minority samples \( \mathcal{D}'_i \) from label \( i \),  
and \( \mathcal{D}'_{i'} \) from label \( i' \) 
using the processes described in \S~\ref{3.1} and \ref{3.2}. We then retrain the ERM model on the updated dataset, which consists of \( \mathcal{D}_{\text{tr}}, \mathcal{D}'_i, \) and \( \mathcal{D}'_{i'} \). 

In each training batch, we separate samples \( \mathcal{B}'_i \in \mathcal{D}'_i \) and \( \mathcal{B}'_{i'} \in \mathcal{D}'_{i'} \). The loss function we employ for each batch is given below:
\begin{eqnarray}
L_{total} &= L_{train} + \gamma_1 L_{gen1} + \gamma_2L_{gen2}~~~\text{s.t.} \\
    L_{gen1} &= \frac{1}{|B'_i|} \sum_{(x,y) \in B'_i} l(f_{\theta}(x), y), \\ 
    L_{gen2} &= \frac{1}{|B'_{i'}|} \sum_{(x,y) \in B'_{i'}} l(f_{\theta}(x), y'). 
\end{eqnarray}
where, \( l \) denotes the  cross-entropy loss, and $L_{train}$ is the original training loss, \( \gamma_1 \) and \( \gamma_2 \) are used for upweighting the new samples. We also provide the algorithm of our overall technique in the supplementary material. 

\vspace{-2mm}
\section{Experiments}
\vspace{-1mm}

\begin{table*}[t!]
\caption{Comparison of worst-group accuracy and average accuracy across three datasets. The \textit{Group Info} column indicates whether group labels are used, where \cmark\cmark\ denotes that validation group labels are also utilized during training. The table is divided into three sections: the first includes methods that address spurious correlations without data augmentation, while the second includes methods that incorporate data augmentation. Our results are in the third section. Where available, we report the mean and standard deviation over three runs in the retraining phase. The best results for each metric are highlighted in bold.
}
\label{tab:method_comparison}
\centering
\small
\begin{tabular}{lccccccc}
\toprule
& {Group‌~Info}& \multicolumn{2}{c}{Waterbirds} & \multicolumn{2}{c}{CelebA} & \multicolumn{2}{c}{MetaShift}\\ 
\cmidrule(lr){3-4} \cmidrule(lr){5-6} \cmidrule(lr){7-8}
Method  & train/val &Worst & Average & Worst & Average & Worst & Average  \\ 
\midrule
DFR \cite{kirichenko2023layerretrainingsufficientrobustness}& \xmark/\cmark\cmark & $92.9 _{\pm 0.2}$ & $93.3 _{\pm 0.5}$ & $88.3 _{\pm 1.1}$ & $91.3 _{\pm 0.3}$ & $72.8 _{\pm 0.6} $ & $77.5_{\pm 0.6}$ \\
JTT \cite{liu2021justtraintwiceimproving} & \xmark/\cmark & $86.7$ & $93.3$ & $81.1$ & $88.0$ & $64.6 _{\pm 2.3}$ & $74.4 _{\pm 0.6}$\\
AFR \cite{qiu2023simplefastgrouprobustness} & \xmark/\cmark & $90.4_{\pm 1.1}$ & $94.2_{\pm 1.2}$ & $82.1_{\pm 0.5}$ & $91.3_{\pm 0.3}$ & - & -\\
MaskTune \cite{taghanaki2022masktunemitigatingspuriouscorrelations} & \xmark/\xmark & $86.4 _{\pm 1.9}$ & $93.0 _{\pm 0.7}$ & $78.0_{\pm 1.2}$ & $91.3_{\pm 0.1}$ & $66.3 _{\pm 6.3}$ & $73.1 _{\pm 2.2}$ \\
\midrule

LISA \cite{yao2022improvingoutofdistributionrobustnessselective} & \cmark/\cmark & $89.2 _{\pm 0.6}$ & $91.8 _{\pm 0.3}$ & $\boldsymbol{89.3} _{\pm 1.1}$ & $92.4 _{\pm 0.4}$ & $59.8 _{\pm 2.3}$ & $70.0 _{\pm 0.7}$\\

DISC \cite{wu2023discovercureconceptawaremitigation} & \xmark/\xmark & $88.7  _{\pm 0.4}$ & $93.8 _{\pm 0.7}$ & - & - & $73.5_{\pm 1.4}$ & $75.5_{\pm 1.1}$\\

DaC \cite{noohdani2024decomposeandcomposecompositionalapproachmitigating} & \xmark/\cmark & $92.3 _{\pm 0.4}$ & $\boldsymbol{95.3} _{\pm 0.4}$ & $81.9 _{\pm 0.7}$& $91.4 _{\pm 1.1}$ & $78.3 _{\pm 1.6}$ & $79.3 _{\pm 0.1}$ \\
\midrule
Base (ERM) & \xmark/\xmark & $74.6$ & $90.2$& $42.2$ &$\boldsymbol{96.0}$ & $67.0$  & $73.9$  \\ 
DBB (ours) & \xmark/\cmark & $\boldsymbol{93.0}_{\pm 0.1}$ & $93.6_{\pm 0.1}$& $85.8 _{\pm 1.4}$ & $87.3_{\pm 0.7}$ & $\boldsymbol{81.2}  _{\pm 0.2}$  & $\boldsymbol{81.3}_{\pm 0.2}$  \\ 
\bottomrule
\end{tabular}
\vspace{-1mm}
\end{table*}

We evaluate our approach, Diffusion-Driven Balancing (DDB), on the standard vision benchmarks designed to assess spurious correlations and distribution shifts, comparing its performance against relevant baselines. Additionally, we conduct an ablation study to examine the impact of the pruning process and hyper-parameter selection.

\subsection{Evaluation Settings}
\label{sec:settings}
\vspace{-1mm}
\textbf{Datasets}. We evaluate our technique on three standard datasets, namely Waterbirds~\cite{article, sagawa2020distributionallyrobustneuralnetworks}, CelebA~\cite{liu2015deeplearningfaceattributes} and MetaShift~\cite{liang2022metashiftdatasetdatasetsevaluating}.  We adopt the standard splits and evaluation protocols~\cite{noohdani2024decomposeandcomposecompositionalapproachmitigating}, \cite{yao2022improvingoutofdistributionrobustnessselective}, details of which are provided in the supplementary material. 
Below, we only focus on the key aspect of spurious correlations found in these  datasets.

\vspace{0.5mm}
\noindent \textit{Waterbirds} \cite{article, sagawa2020distributionallyrobustneuralnetworks}: 
This dataset is used for vision classification  with sample labelled as ``landbirds" and ``waterbirds." The labels are spuriously correlated with the scene background, as most landbirds appear in land backgrounds, and most waterbirds appear in water backgrounds. 

\vspace{0.5mm}
\noindent\textit{CelebA} \cite{liu2015deeplearningfaceattributes}: In this dataset, the task is to classify celebraties as ``blond",``not blond". The spurious correltation exists between the gender of the person and their hair color as most females in the dataset have blond hair.


\vspace{0.5mm}
\noindent\textit{MetaShift} \cite{liang2022metashiftdatasetdatasetsevaluating}: Our setup for this dataset follows \cite{wu2023discovercureconceptawaremitigation}. It is a classification dataset with labels ``dogs" and ``cats." In the training set, the ``dogs" label is spuriously correlated with background objects such as benches and bikes, while the ``cats" label is correlated with objects such as beds and sofas. This correlation does not exist in the test set.

\vspace{0.5mm}
\noindent\textbf{Compared Methods}. We compared DDB with the existing state-of-the-art methods that address spurious correlations, both with and without data augmentation. Specifically, we perform comparisons with DFR \cite{kirichenko2023layerretrainingsufficientrobustness}, JTT \cite{liu2021justtraintwiceimproving}, MaskTune \cite{taghanaki2022masktunemitigatingspuriouscorrelations} and AFR \cite{qiu2023simplefastgrouprobustness} for methods without data augmentation. For methods incorporating data augmentation, we compared with DISC \cite{wu2023discovercureconceptawaremitigation}, LISA \cite{yao2022improvingoutofdistributionrobustnessselective}, and DAC~\cite{noohdani2024decomposeandcomposecompositionalapproachmitigating}. For more details about these methods, see \S~\ref{sec:related}.

\vspace{0.5mm}
\noindent\textbf{Models}. For a fair comparison, similar to other compared methods, we use a ResNet-50 model pre-trained on ImageNet as the classifier. In our approach, for textual inversion and sample generation, we leveraged the Hugging Face implementation of Stable Diffusion \cite{rombach2022highresolutionimagesynthesislatent}\footnote{\url{https://huggingface.co/stabilityai/stable-diffusion-2-inpainting}}. Additionally, for the pruning phase, we employed the PyTorch implementation of Integrated Gradients \cite{sundararajan2017axiomaticattributiondeepnetworks}\footnote{\url{https://pytorch.org/tutorials/beginner/introyt/captumyt.html}}.

\vspace{0.5mm}
\noindent\textbf{Hyperparameters.} For all datasets, hyperparameters were tuned to maximize worst-group accuracy on the validation set. We selected the lowest-loss $K$ portion of samples per class and the upweighting factor $\gamma$ from $\{0, 0.1, 0.2, \dots, 1\}$ and $\{1, 2, 3, \dots, 10\}$, respectively. The threshold $\mathrm{P}$ was determined using Integrated Gradients (IG) scores from a subset of generated samples.

For textual inversion, we selected 20–40 minority-group samples per class, either from the validation set using group labels or from the training set based on high-loss samples. We provide further details on this in  \S~\ref{ablation} and supplementary material. It is noteworthy that our technique does not require group labeling for the training set.

\vspace{-1mm}
\subsection{Results}
\vspace{-1mm}
The results of DFR \cite{kirichenko2023layerretrainingsufficientrobustness}, JTT \cite{liu2021justtraintwiceimproving}, AFR \cite{qiu2023simplefastgrouprobustness}, and MaskTune \cite{taghanaki2022masktunemitigatingspuriouscorrelations}, which enhance group robustness without data augmentation, as well as LISA \cite{yao2022improvingoutofdistributionrobustnessselective}, DISC \cite{wu2023discovercureconceptawaremitigation}, and DaC \cite{noohdani2024decomposeandcomposecompositionalapproachmitigating}, which incorporate data augmentation, along with the ERM baseline, are presented in Table \ref{tab:method_comparison}. We report worst-case and average accuracy from the respective papers for the Waterbirds and CelebA datasets, while MetaShift results are taken from \cite{noohdani2024decomposeandcomposecompositionalapproachmitigating}. These sources follow the same evaluation protocols as followed in our experiments. 

Among the methods that do not rely on data augmentation, our approach achieves superior worst-group accuracy—a key metric for group robustness—on Waterbirds and MetaShift. The MetaShift dataset, which involves domain shift, further highlights DDB’s effectiveness in addressing such shifts by redirecting the ERM model’s attention to causal features. Unlike existing methods, our approach generates rare samples to enhance robustness, a capability absent in this category. On CelebA, our results remain competitive with state-of-the-art methods.

For methods leveraging data augmentation, DDB also achieves superior worst-group accuracy on Waterbirds and MetaShift. LISA requires group labels for training, whereas our method operates without them while achieving comparable performance on CelebA. Similar to our approach, DaC addresses spurious correlations compositionally by combining causal and non-causal image components. However, DaC primarily mitigates spurious objects but remains vulnerable to spurious features. CelebA, which involves spurious feature correlations (e.g., hair length and label), presents a challenge for DaC, whereas DDB remains robust to both spurious objects and features, delivering competitive performance with state-of-the-art methods. For further details on different types of spurious correlations, we refer to the supplementary material.

\begin{table}[t!]
\caption{Results for the pruning process. The first column indicates the number of selected samples in each class for modification. The \textit{Pruned} column shows the number of samples removed after the modification based on the pruning process. The last column reports the worst-group accuracy of our approach without pruning. These results can be compared with the last row of Table~\ref{tab:method_comparison} to see the impact of pruning. 
}\vspace{-2mm}
\label{tab:withoutprun}
\centering
\small
\begin{tabular}{c|cccc}
\toprule
 Dataset & \multicolumn{2}{c}{Number of samples} & Pruned & DDB - Pruning\\ 
\midrule
 \multirow{2}{4em}{Waterbirds} & Landbird& 1300 & 531 &  \multirow{2}{4em}{91.28}  \\
  & Waterbird & 1112 & 393 & \\
\midrule
\multirow{2}{3em}{CelebA}  & NotBlond& 120000 & 7439 & \multirow{2}{4em}{81.7}  \\
 & Blond & 2000 & 869 & \\ 
 \midrule
  \multirow{2}{4em}{MetaShift} & Cat &  400 & 134 &  \multirow{2}{4em}{80.6}  \\
 & Dog & 400 & 299 & \\
\bottomrule
\end{tabular}
\vspace{-3mm}
\end{table}

\vspace{-1mm}
\subsection{Ablation Study}
\label{ablation}
\vspace{-1mm}

\textbf{Pruning.} \label{pruning} As shown in Table \ref{tab:withoutprun}, pruning operation enhances the performance of our approach. Inspecting the results, in the MetaShift dataset, $75\%$ of newly generated samples for the \textit{Cat} class (i.e., \textit{Dog} samples altered to \textit{Cat}) are pruned. This is due to many \textit{Dog} class images containing dogs that are either too small or entirely absent. Consequently, Stable Diffusion struggles to modify these regions effectively, leading to corrupted samples that require pruning. We discuss this more in the supplementary material.

\begin{table}[t]
\caption{Worst-group accuracy change with different number of minority groups (NonBlond Female and Blond Male) added to the CelebA training set.}\vspace{-2mm}

\label{tab:addedsamples}
\centering
\small
\begin{tabular}{c|ccc}
\toprule 
NonBlond&\multicolumn{3}{c}{Blond Male} \\
Female&2000&3000&4000\\
\midrule

1000 & 85.6 & 86.6 & 85.6  \\
\midrule
2000 & 87.1 & 86.8 & 86.7\\
\bottomrule
\end{tabular}
\vspace{-1mm}
\end{table}

\begin{table}[t]
\caption{Number of  samples selected to modify  each group.
} \vspace{-3mm}
\label{tab:numbersample}
\centering
\small
\begin{tabular}{c|cccc}
\toprule

 \multirow{2}{3em}{Dataset}  & \multicolumn{2}{c}{label = 0} & \multicolumn{2}{c}{label = 1}\\
 & Majority& Minority &Majority& Minority \\

\midrule
Waterbirds & 1292 & 8 & 1057 & 56 \\
\midrule
CelebA & 6875 & 5125 & 1999& 1
\\
\midrule
Metashift & 256 & 144 & 290 & 110
\\
\bottomrule
\end{tabular}
\vspace{-3mm}
\end{table}

Similarly, a considerable number of the modified \textit{NotBlond} images get  pruned in Table \ref{tab:withoutprun}. On inspection, we discover that many samples in this class feature bald individuals or those wearing hats, making it challenging for LangSAM to predict masks accurately. We also observed that LangSAM sometimes struggled with reliable hair region identification, leading to corrupted generated images. These results highlight the need of pruning to substantiate the efficacy of our method. We provide more details in the supplementary material for failure types.

\vspace{0.5mm}
\noindent\textbf{\textcolor{black}{Number of Selected Samples.}}  
\label{sec:numbersec}  
As shown in Table \ref{tab:addedsamples}, adding samples from the minority group to the dataset generally enhances the robustness of the ERM model. However, the optimal performance is achieved when an appropriate number of samples is added to the dataset. The ideal number is simply a function of the imbalance level of the dataset and the number of samples required to achieve the balance. 
In Table \ref{tab:numbersample}, our chosen  number of samples for the desired modification are reported. Increasing the number of  samples beyond these raises the likelihood of selecting the minority group samples. In the CelebA dataset, the number of selected samples for label 0 is nearly balanced between the majority and minority groups. This is because CelebA does not exhibit a clear majority-minority structure across both classes. Specifically, the number of non-blond males and non-blond females in the training dataset is almost equal. Hence, incorporating group labels helps mitigate the issue and enhances the effectiveness of the approach. In Metashift and Waterbirds dataset, we simply use the number of samples necessary to achieve the balance. 

\vspace{0.5mm}
\noindent\textbf{Textual Inversion.}  
To capture the causal features of each class, we fine-tune the text embedding model by optimizing token embeddings through minimizing  the loss given in Eq.~(\ref{eq:TokenInversion}) on a small dataset. Table \ref{tab:inversion} presents the number of selected samples in this minimization and their impact on the performance of our method. The first results column corresponds to our approach without textual inversion. The later columns show results with the inversion of the samples for Dataset Size $>$ 0. As demonstrated, incorporating textual inversion enhances performance.

As identified by Table \ref{tab:inversion}, poor textual inversion, which leads to generation of more undesirable samples, gets treated by our method (automatically)  with a higher number of pruned samples. The largest number of pruned samples occurs when textual inversion is not employed, whereas the lowest number is observed in the setting that achieves the highest worst-group accuracy, indicating improved quality in the generated images.
Hence, to mitigate inaccuracies introduced by the LangSAM model, we prioritize sample selection from minority groups. This mitigates the case of inclusion of spurious region inclusion in the masks, which can potentially lead Stable Diffusion to propagate these artifacts into the generated images because it attempts to generate a minority-group sample using the learned token. We also provide further details on the textual inversion process  in the supplementary material.

\begin{table}[t!]
\caption{Number of (automatically) pruned samples and worst-group accuracy (WGA) under different textual inversion settings for  the Waterbirds dataset. The first row indicates the number of samples selected for textual inversion. The first results column represents our approach without textual inversion.  
} \vspace{-1mm}
\label{tab:inversion}
\centering
\small
\begin{tabular}{c|ccccc}
\toprule

Dataset Size & 0 & 10 &20& 30& 40 \\

\midrule
Pruned & 1018 & 818 & 804 & 737 & 782 \\
\midrule
WGA & 88.9 & 90.5 & 89.9& 93 & 90.3
\\
\bottomrule
\end{tabular}
\vspace{-4mm}
\end{table}

\vspace{-2mm}
\section{Conclusion}
\vspace{-1.5mm}
We introduced Diffusion Driven Balancing (DDB), a novel approach for generating underrepresented samples in training data using text-to-image generative models. We leveraged diffusion models to selectively modify specific regions of images in a controlled manner to generate scarce samples, and mitigate spurious correlations.
We employed textual inversion for this purpose. We also introduced a pruning mechanism that leverages attribution scores and ERM predictions to filter out undesired generative  samples. Our  results demonstrate that the proposed method effectively mitigates spurious correlations and enhances classification performance under distribution shifts. Although our primary exploration was about mitigating  spurious correlations, which is shown by strong performance across a variety of datasets, our method has the potential to address general data scarcity problem. 

\noindent\textbf{Limitations and Ethical Considerations.}  
Our method leverages existing diffusion and  language-segmentation models. This potentially limits the applicability of our method as per the abilities of these models.
However, with the rapid advances in generative and language models, this limitation is of minor nature. Continuously growing abilities of these models will proportionally strengthen our underlying framework.   
We also note that since our method provides precise automated control for image modification, it poses a potential risk of misuse in Deepfake generation. Hence, we advocate only a responsible use of our method.

\section{Acknowledgment}
\vspace{-1mm}
Naveed Akhtar is a recipient of the Australian Research Council Discovery Early Career Researcher Award (project \# DE230101058) funded by the Australian Government. This work is also partially supported by Google Research Scholar Program Award.

{
    \small
    \bibliographystyle{ieeenat_fullname}
    \bibliography{main}
}
\clearpage
\setcounter{page}{1}
\maketitlesupplementary

\section{Dataset Details}
\textbf{Waterbirds}. This dataset is introduced by \cite{sagawa2020distributionallyrobustneuralnetworks} and is a synthesized collection created by cropping birds from the CUB dataset \cite{article} and pasting them onto backgrounds from the Places dataset \cite{zhou2016placesimagedatabasedeep}. In this dataset, seabirds and waterfowl are labeled as ``Waterbirds,'' while other birds are labeled as ``Landbirds.'' The backgrounds are categorized into two types: ocean and natural lake for the water background, and bamboo forest and broadleaf forest for the land background. 

In the dataset, waterbirds with a water background and landbirds with a land background are considered the majority groups, while waterbirds with a land background and landbirds with a water background are considered the minority groups. The number of samples in each group is shown in Table \ref{tab:waterbirds_samples}.

\textbf{CelebA}. Following \cite{sagawa2020distributionallyrobustneuralnetworks}, we utilized the "blond" and "non-blond" attributes from celebrity images in \cite{liu2015deeplearningfaceattributes} as labels. In this dataset, there is a spurious correlation between gender and label. The majority groups are typically considered to be blond females and non-blond males, while the minority groups are considered to be blond males and non-blond females. However, as mentioned in the paper, the number of non-blond females and males are almost equal, and thus, this dataset only have one minority group. We also used the class-balanced dataset for our approach, similar to \cite{noohdani2024decomposeandcomposecompositionalapproachmitigating}. The number of samples in each group is shown in Table \ref{tab:celebA_samples}.

\textbf{Metashift}. This dataset was introduced by \cite{liang2022metashiftdatasetdatasetsevaluating} for a binary classification task with labels "dog" and "cat." The dataset exhibits a diversity shift, as categorized by \cite{ye2022oodbenchquantifyingunderstandingdimensions}, where test images contain different spurious attributes compared to the training set. Following \cite{wu2023discovercureconceptawaremitigation}, cat images in the training set have spurious attributes, such as "bed" or "sofa" in the background, while dog images have spurious attributes, such as "bench" and "bike." In contrast, the test images do not contain these attributes; instead, the spurious attribute in the test images is "shelf." The number of samples in each group is shown in Table \ref{tab:metashift_samples}.

\begin{table}[ht!]
\caption{The number of samples in the training, validation, and test sets for the Waterbirds dataset}
\label{tab:waterbirds_samples}
\centering
\small
\begin{tabular}{c|ccc}
\toprule

Split & Train & Validation & Test \\
\midrule
(landbird, land background) & 3,498 & 467 & 2,255\\
\midrule
(landbird, water background) & 184 & 466 & 2,255 \\ 
\midrule
(waterbird, land background) & 56 & 133 & 642 \\
\midrule
(waterbird, water background) & 1,018 & 133 & 642   
\end{tabular}
\end{table}

\begin{table}[ht!]
\caption{The number of samples in the training, validation, and test sets for the CelebA dataset}
\label{tab:celebA_samples}
\centering
\small
\begin{tabular}{c|ccc}
\toprule

Split & Train & Validation & Test \\
\midrule
(NonBlond, female) & 12,426 & 8,535 & 9,767\\
\midrule
(NonBlond, male) & 11,841 & 8,276 & 7,535 \\ 
\midrule
(Blond, female) & 22,880 & 2,874 & 2,480 \\
\midrule
(Blond, male) & 1,387 & 182 &  180  
\end{tabular}
\end{table}

\begin{table}[ht!]
\caption{The number of samples in the training, validation, and test sets for the Metashift dataset}
\label{tab:metashift_samples}
\centering
\small
\begin{tabular}{c|ccc}
\toprule

Split & Train & Validation & Test \\
\midrule
(Cat, sofa) & 231 & 0 & 0\\
\midrule
(Cat, bed) & 380 & 0 & 0\\ 
\midrule
(Dog, bench) & 145 & 0 &  0\\
\midrule
(Dog, bike) & 367 & 0 &  0  \\
\midrule
(Cat, shelf) & 0 & 34 & 201   \\
(Dog, shelf) & 0 & 47 &  259  \\
\end{tabular}
\end{table}
The time required to generate new samples is shown in Table \ref{tab:gentime} using an A100 GPU. This includes the masking process, generation using the diffusion model, and the computation of the Integrated Gradient (IG) score.

\begin{table}[ht!]
\caption{Generation time in datasets.}
\label{tab:gentime}
\centering
\small
\begin{tabular}{c|ccc}
\toprule

 Generation time & Metashift & CelebA & Waterbirds \\
 (hours) & 0.5 & 5 & 1\\

\end{tabular}
\end{table}
\subsection{Details on CelebA and Spurious Features}

\begin{figure}[t]
    \centering
    \includegraphics[width=0.98\linewidth]{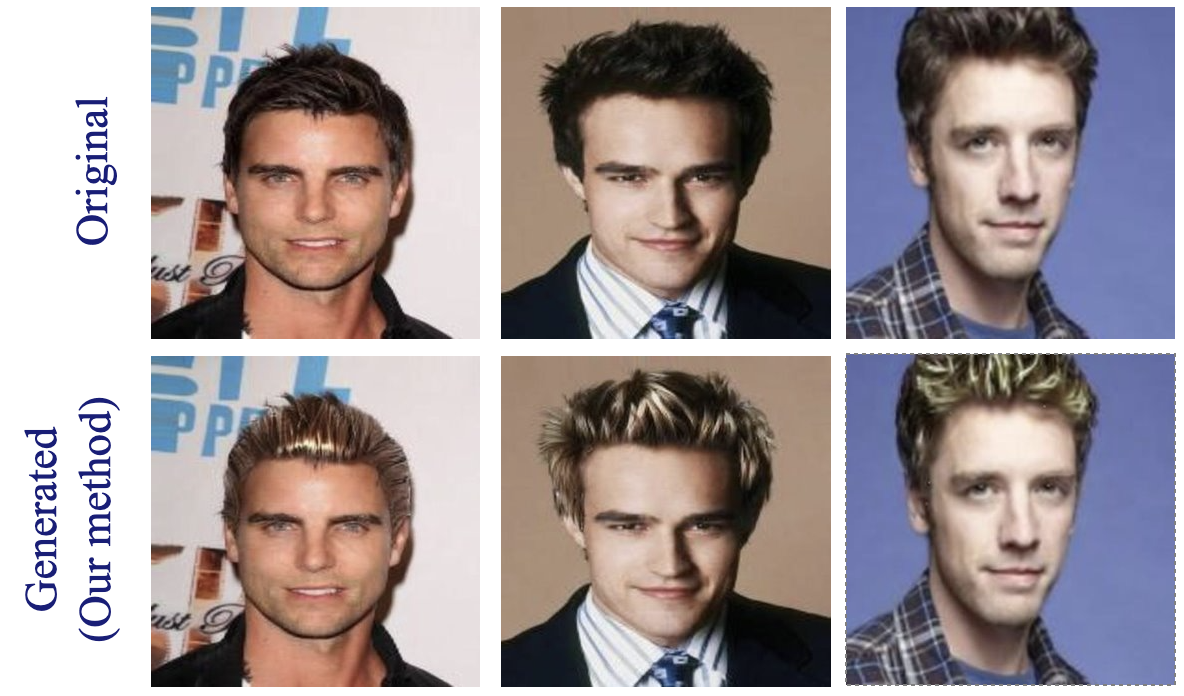}
    \caption{Generated samples in the CelebA dataset. The generated samples exhibit the spurious features of the non-blond class, as they are short and not wavy.}
    \label{fig:spurious_example}
\end{figure}

In addition to spurious objects (such as backgrounds in the Waterbirds dataset or gender-based facial attributes in the CelebA dataset), spurious features can also exist within datasets. In the CelebA dataset, there is a correlation between hair color and other hair attributes, such as hair length or shape. For example, there is an imbalance between long blond hair and long non-blond hair, as well as between wavy blond hair and wavy non-blond hair \cite{noohdani2024decomposeandcomposecompositionalapproachmitigating}. Our approach makes the ERM model robust to both of these spurious correlations by balancing the dataset in both ways. In Figure \ref{fig:spurious_example}, we show multiple images that are not wavy and have short blond hair, addressing this type of spurious correlation.

\section{Experiments}
\subsection{DDB Training}

\textbf{ERM Model}. We utilized the ERM model for the CelebA and Waterbirds datasets from \cite{noohdani2024decomposeandcomposecompositionalapproachmitigating}, and the Metashift dataset from \cite{wu2023discovercureconceptawaremitigation}.

\textbf{Hyperparameters}. For textual inversion training, we utilized the Huggingface implementation. The detailed hyperparameters for our method on the three datasets are shown in Table \ref{tab:hyperparameters}. 

\begin{table*}[t!]
\caption{Hyperparameters. The batch size and learning rate are for the retraining phase of the ERM model.}
\label{tab:hyperparameters}
\centering
\small
\begin{tabular}{c|cccccc}
\toprule

Dataset & Batch size & Learning rate & $\gamma_1$& $\gamma_2$& $\mathrm{P}_1$& $\mathrm{P}_2$\\

\midrule
CelebA & 64 & 0.00001 & 2 & 7 & 2 & 0.3 \\
\midrule
Metashift & 16 & 0.00005 & 1 & 4 & 0.3 & 2 \\
\midrule
Waterbirds & 32 & 0.000005 & 6 & 10 & 1 & 1\\
\bottomrule
\end{tabular}
\vspace{-3mm}
\end{table*}

\textbf{Algorithm}. A formal algorithm of our method is provided in Algorithm \ref{alg:psudocode_ddb}.

\begin{algorithm}
  \caption{Diffusion-Driven-Balancing (DDB)}
  \label{alg:psudocode_ddb}
  
  \SetAlgoLined
  \KwIn{Training Dataset $\mathcal{D}_{train}$, Class $i$,$i'$, Inpainting Model $SD$, Masking Model $m$, Hyperparametes $K_i,K_{i'}, \gamma_1, \gamma_2, \mathrm{P}_{i}, \mathrm{P}_{i'}$}

  $Textual\_Dset_i \leftarrow \text{Pick a few samples $(x,y) \in \mathcal{D}_{train}$}$\;
  $Textual\_Dset_{i'} \leftarrow \text{Pick a few samples $(x,y') \in \mathcal{D}_{train}$}$\;
  $C^*_i = TextualInversion(SD, Textual\_Dset_i)$\;
  $C^*_{i'} = TextualInversion(SD, Textual\_Dset_{i'})$\;
    $\mathcal{D}_i \leftarrow SaveLowLossSamples(\mathcal{D}_{train}, i, K_i)$\;
    $\mathcal{D}_{i'} \leftarrow SaveLowLossSamples(\mathcal{D}_{train}, i', K_{i'})$ \;
    \tcp{Equation \ref{softmax}}
    $\Psi_i \leftarrow MeanSoftMax(\mathcal{D}_{i},i)$ \;
    $\Psi_i \leftarrow MeanSoftMax(\mathcal{D}_{i'},i')$\;
    $\mathcal{D'}_{i'} = GenerateNewDataset(f,i,i',\Psi_i, \mathrm{P}_{i'}, m, C^*_{i'}, SD)$\;
    $\mathcal{D'}_{i} = GenerateNewDataset(f,i',i,\Psi_i',\mathrm{P}_i, m, C^*_{i}, SD)$\;
    $\mathcal{D}_{train} \leftarrow \mathcal{D}_{train} \cup \mathcal{D'}_i \cup \mathcal{D'}_{i'}$ \;
    \For{epoch = 1,2,3...,O}
    {
    \For{batch $\mathcal{B} in \mathcal{D}_{tr}$}
    {
    $B'_i = Samples from \mathcal{B} \in \mathcal{D'}_i$\;
    $B'_{i'} = Samples from \mathcal{B} \in \mathcal{D'}_{i'}$\;
    $L_{CE} = \frac{1}{|B|} \sum_{(x,y) \in B} l(f_{\theta}(x), y)$\;
    $L_{gen1} = \frac{1}{|B'_i|} \sum_{(x,y) \in B'_i} l(f_{\theta}(x), y)$\;
    $L_{gen2} = \frac{1}{|B'_{i'}|} \sum_{(x,y) \in B'_{i'}} l(f_{\theta}(x), y')$\;
    $L_{total} = L_{CE} + \gamma_1 L_{gen1} + \gamma_2L_{gen2}$\;
    $f \leftarrow UpdateWeights(L_{total})$
    }
    }
\end{algorithm}

 \begin{figure}[t]
    \centering
    \includegraphics[width=\linewidth]{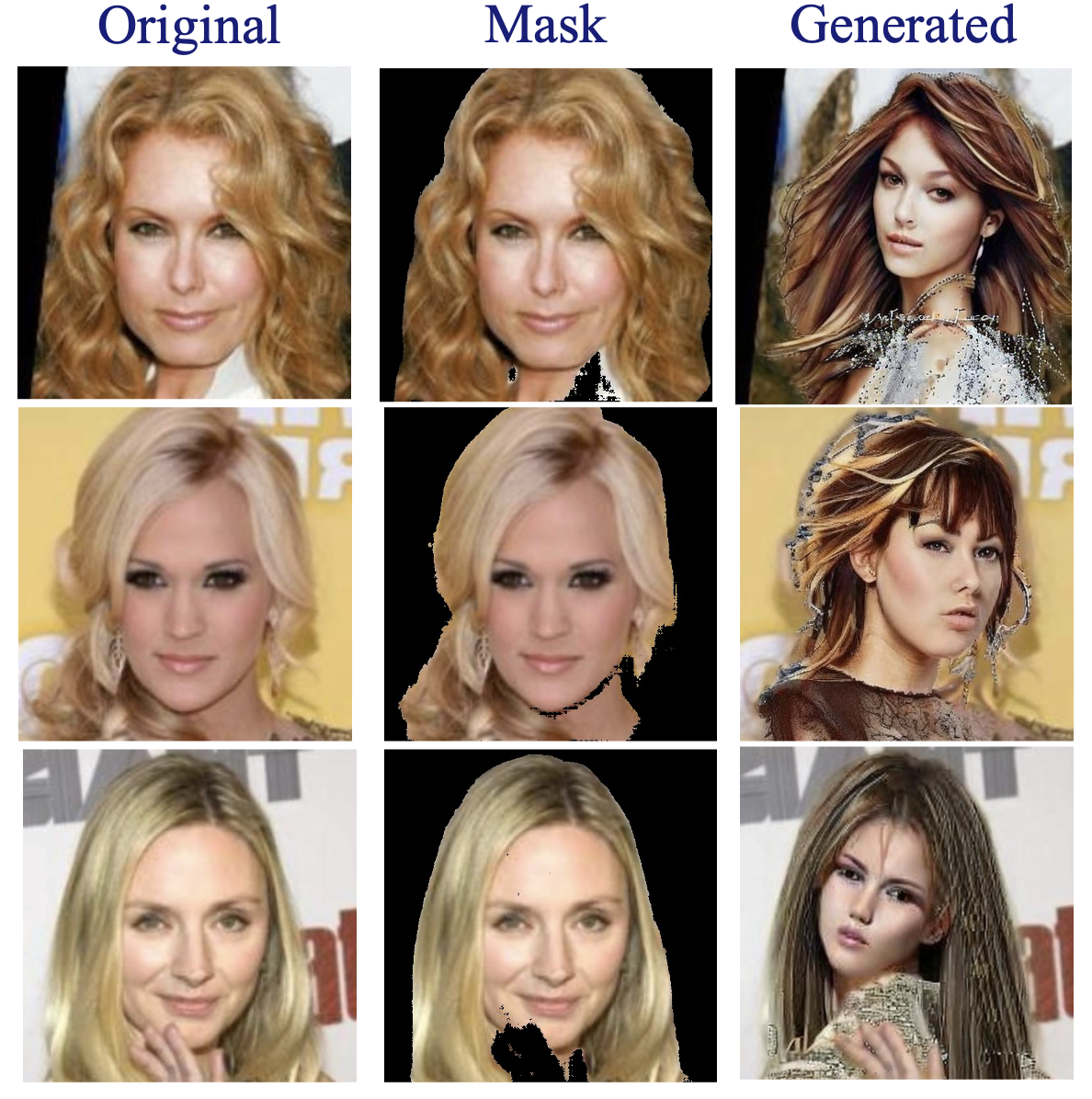}
    \caption{Generated non-blond females in cases where masking fails.}
    \label{fig:nonblondfemale}
\end{figure}

\subsection{Textual inversion}
\label{texsup}
\begin{figure*}[t]
    \centering
    \includegraphics[width=\linewidth]{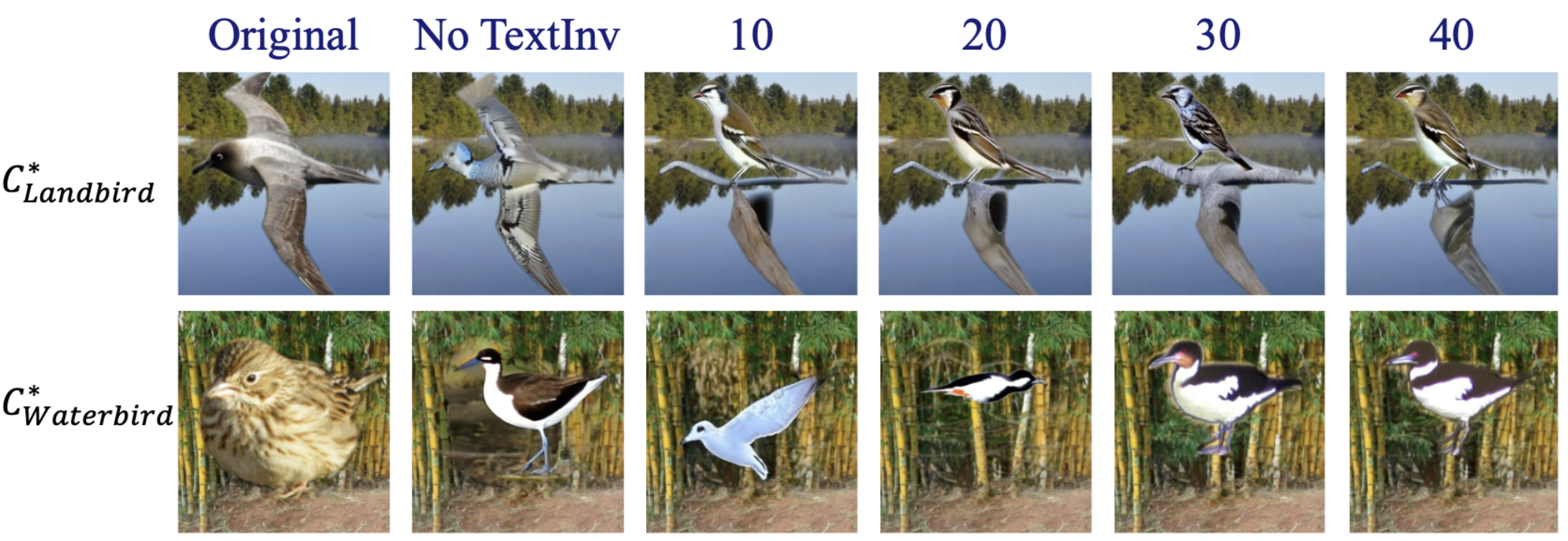}
\caption{Examples of generated samples for landbirds and waterbirds with varying textual inversion dataset sizes, as well as samples generated without utilizing textual inversion and the initial image.}

    \label{fig:texinvers}
\end{figure*}

Figure \ref{fig:texinvers} shows the impact of the textual inversion dataset size. As demonstrated, the optimal number of samples leads to the specified token better representing the causal parts, resulting in higher-quality images.

We selected minority group samples for the textual inversion dataset to ensure that the learnable token does not reconstruct the spurious feature $s_{\text{maj}}$, thereby preventing error propagation when other components fail. For instance, in the CelebA dataset, where the mask is intended to isolate the hair, if the mask encompasses both the face and the hair, Stable Diffusion will regenerate the face in accordance with the learned representation of the minority group samples, provided the token has been trained to capture such features. As shown in Figure \ref{fig:nonblondfemale}, the mask includes the face region as well. However, the learned token generates a non-blond woman, as the text embedding model was trained on a dataset of non-blond women.

\subsection{Pruning}

\begin{figure}[t]
    \centering
    \includegraphics[width=\linewidth]{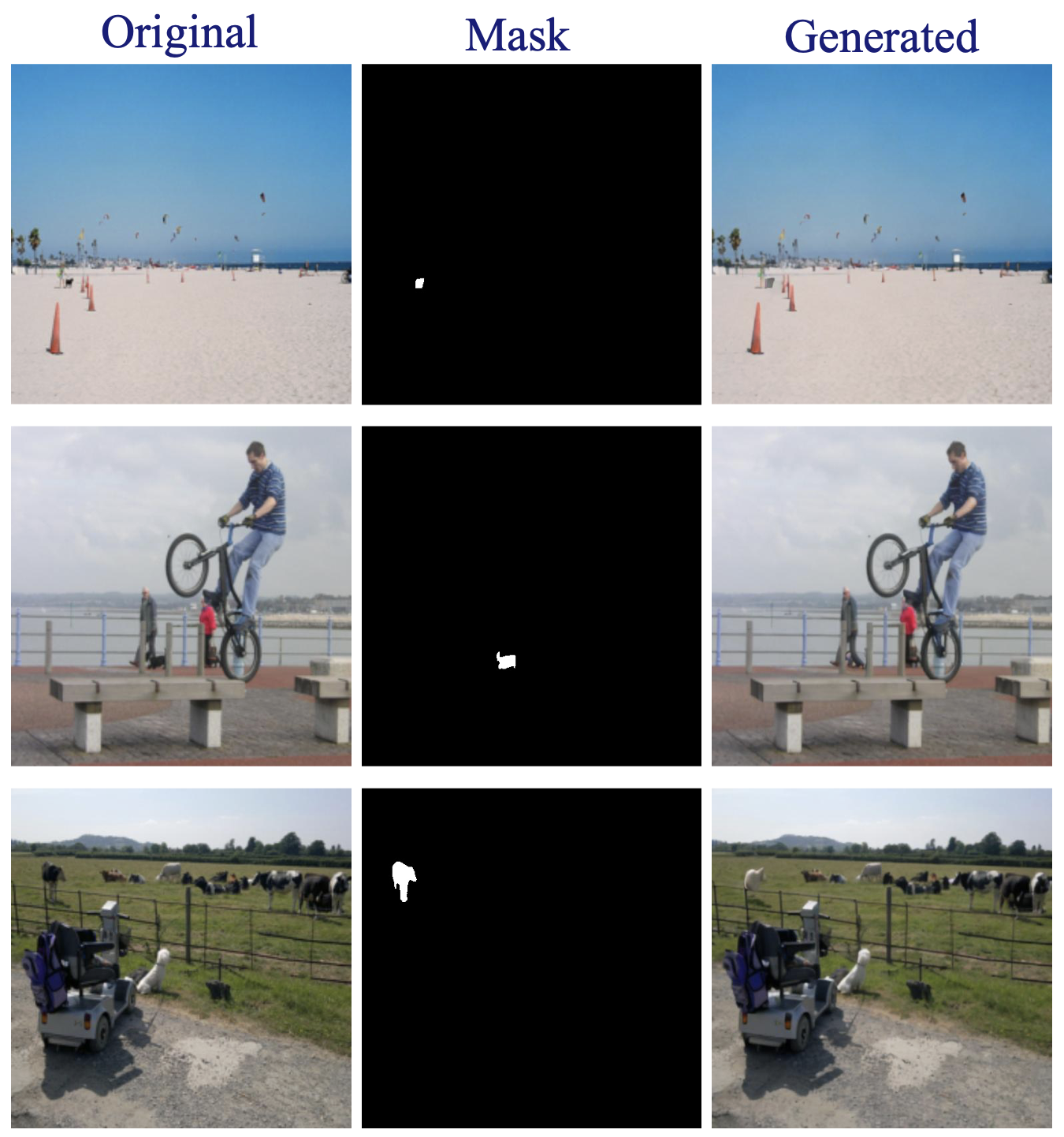}
    \caption{Metashift samples when the stable diffusion cannot perform well because of the initial image.}
    \label{fig:metafailed}
\end{figure}

\begin{figure}[t]
    \centering
    \includegraphics[width=\linewidth]{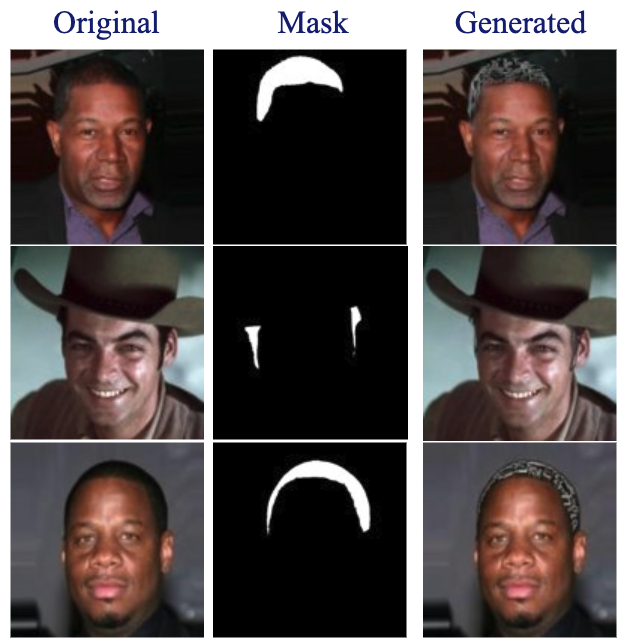}
    \caption{Example of bald samples or samples wearing hat in the CelebA dataset.}
    \label{fig:celebAfailed}
\end{figure}

As mentioned in the paper, in the Metashift dataset, many samples in the dog class are too small compared to the image or other animals present, leading the masking model to mistakenly identify these non-relevant parts as the causal component. Some examples of these problematic samples are shown in Figure \ref{fig:metafailed}. In the CelebA dataset, identifying hair is challenging for the masking model, as shown in Figure \ref{fig:spurious_example}. To address this limitation, we use textual inversion, as explained in \S~\ref{texsup}. Additionally, many bald samples wearing hats exist in the non-blond class, which results in corrupted generated images that need to be pruned. An example of this case is shown in Figure \ref{fig:celebAfailed}.

\subsection{Imagenet-A}

\begin{figure}[t]
    \centering
    \includegraphics[width=\linewidth]{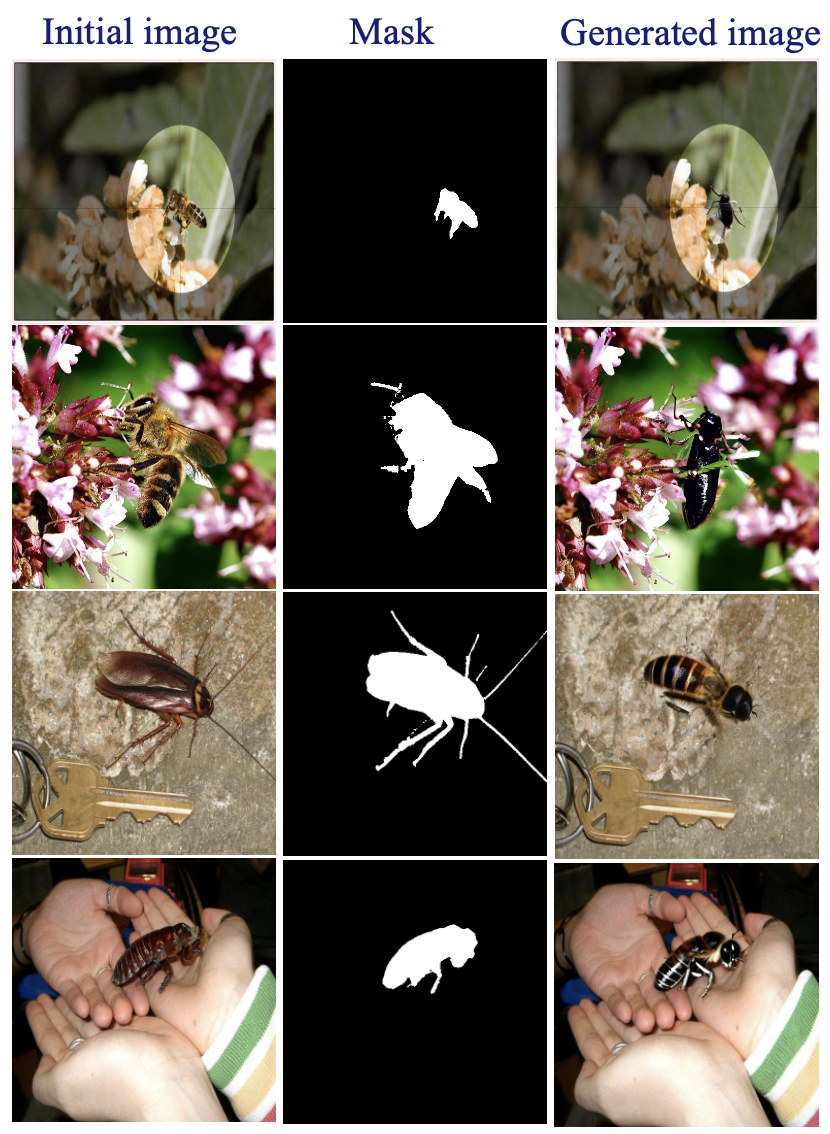}
    \caption{Examples of generated samples along with the corresponding masks and initial images. The generated images belong to the cockroach class, with initial images taken from the bee class, and vice versa.}
\label{fig:Imagenet}
\end{figure}

For further experiments and to evaluate DDB on more general tasks, we assessed our approach on two classes from ImageNet-A~\cite{hendrycks2021naturaladversarialexamples}—natural adversarial examples: \textit{Cockroach} and \textit{Bee}. This dataset contains samples on which standard machine learning models typically perform poorly. Our method improved the average accuracy of the ERM baseline on this dataset from $43.26\%$ to $62.58\%$. We used ImageNet-1k~\cite{5206848} as the source of training samples. The ERM model was trained using the SGD optimizer with a learning rate of $0.001$, while the retraining phase used the Adam optimizer with a learning rate of $0.0005$. The batch size was set to $32$ for both stages. We also set the hyperparameters $\gamma_1 = 1$ and $\gamma_2 = 1$ in this experiment. Qualitative images are available in Fig. \ref{fig:Imagenet}.

\subsection{Qualitative images}
\begin{figure}[t]
    \centering
    \includegraphics[width=\linewidth]{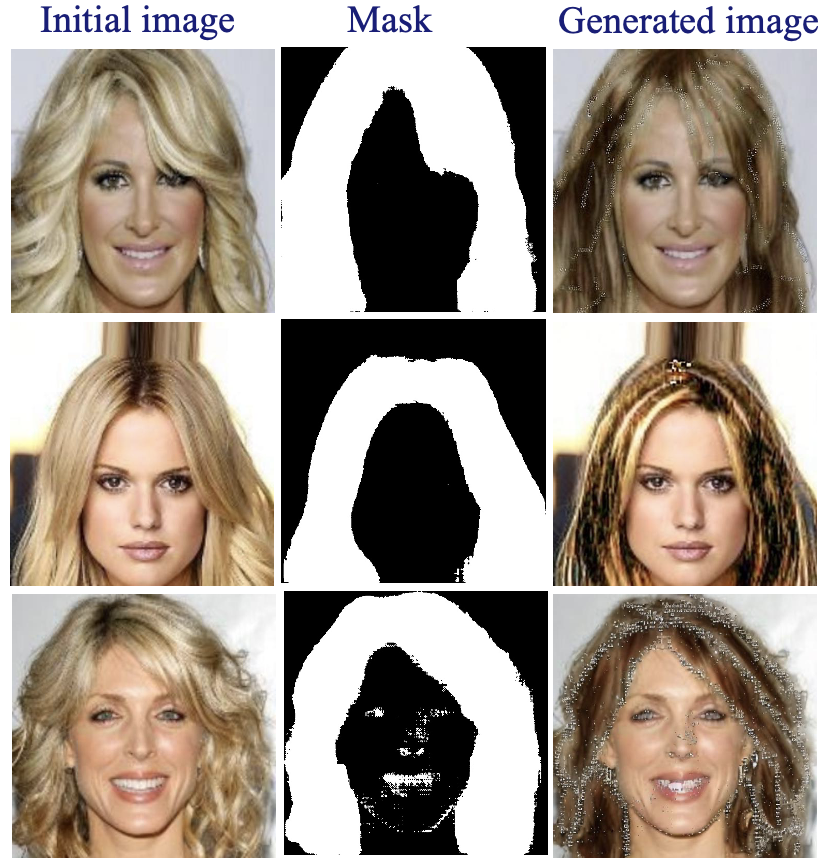}
    \caption{Examples of generated female samples along with the corresponding masks and initial images. The initial images are belong to male class.}
\label{fig:femalenew}
\end{figure}

\begin{figure}[t]
    \centering
    \includegraphics[width=\linewidth]{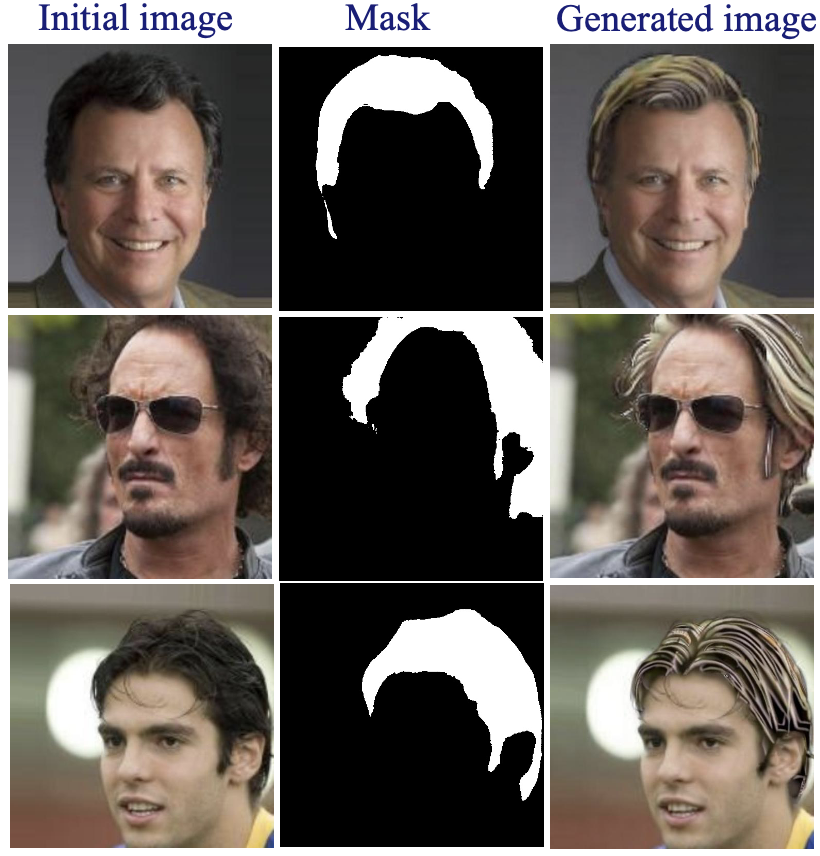}
    \caption{Examples of generated male samples along with the corresponding masks and initial images. The initial images are belong to female class.}
\label{fig:malenew}
\end{figure}

\begin{figure}[t]
    \centering
    \includegraphics[width=\linewidth]{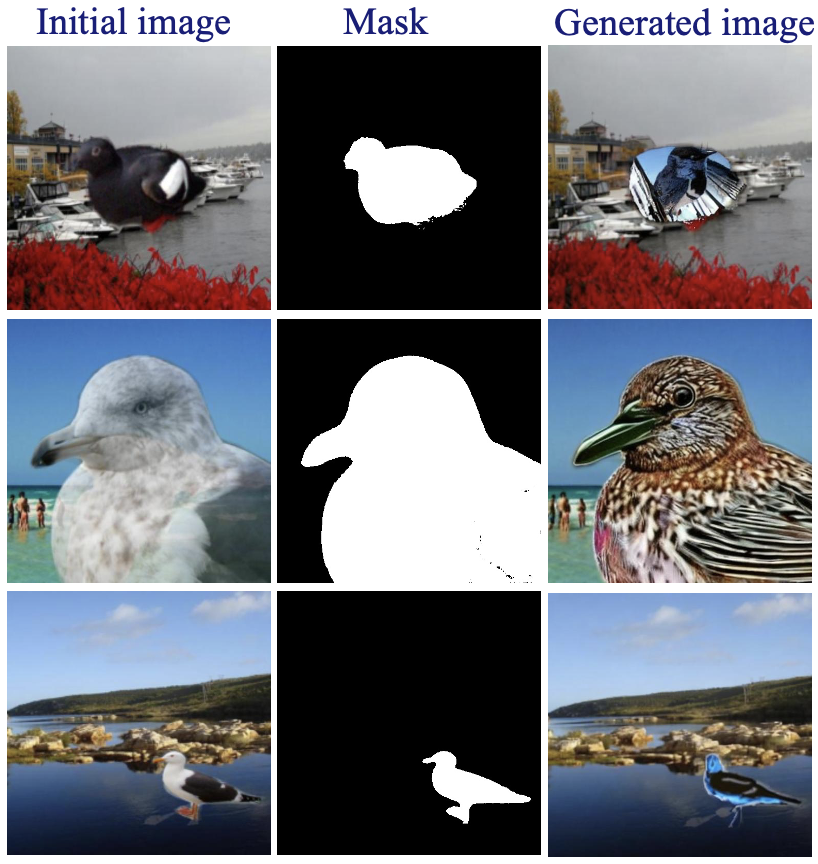}
    \caption{Examples of generated landbird samples along with the corresponding masks and initial images. The initial images are belong to waterbird class.}
\label{fig:landbirdnew}
\end{figure}

\begin{figure}[t]
    \centering
    \includegraphics[width=\linewidth]{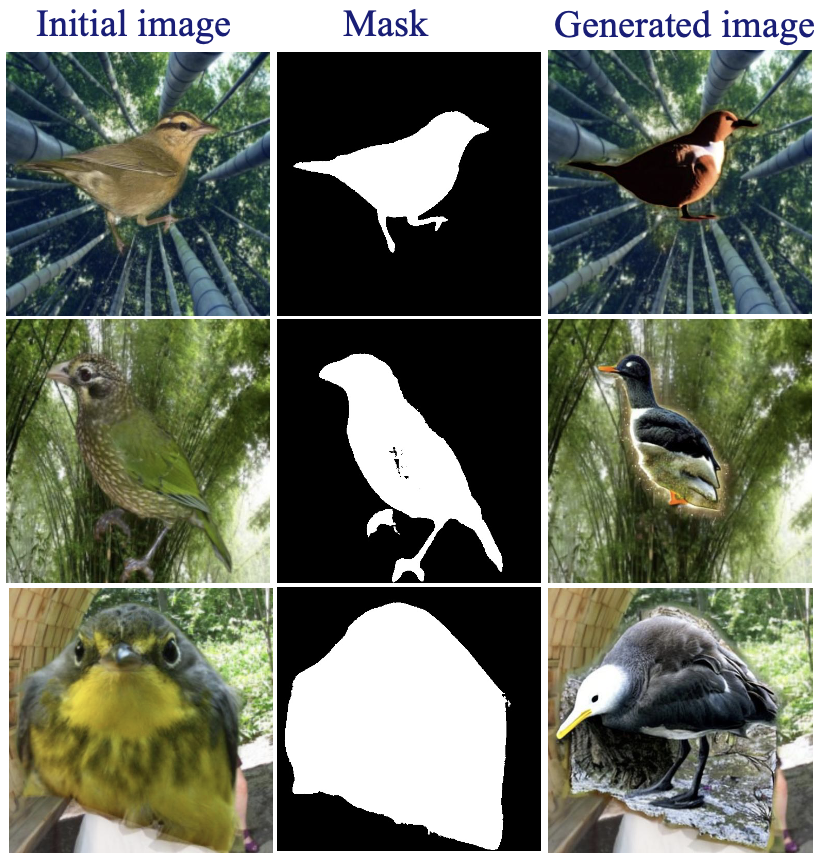}
    \caption{Examples of generated waterbird samples along with the corresponding masks and initial images. The initial images are belong to landbird class.}
\label{fig:waterbirdnew}
\end{figure}

\begin{figure}[t]
    \centering
    \includegraphics[width=\linewidth]{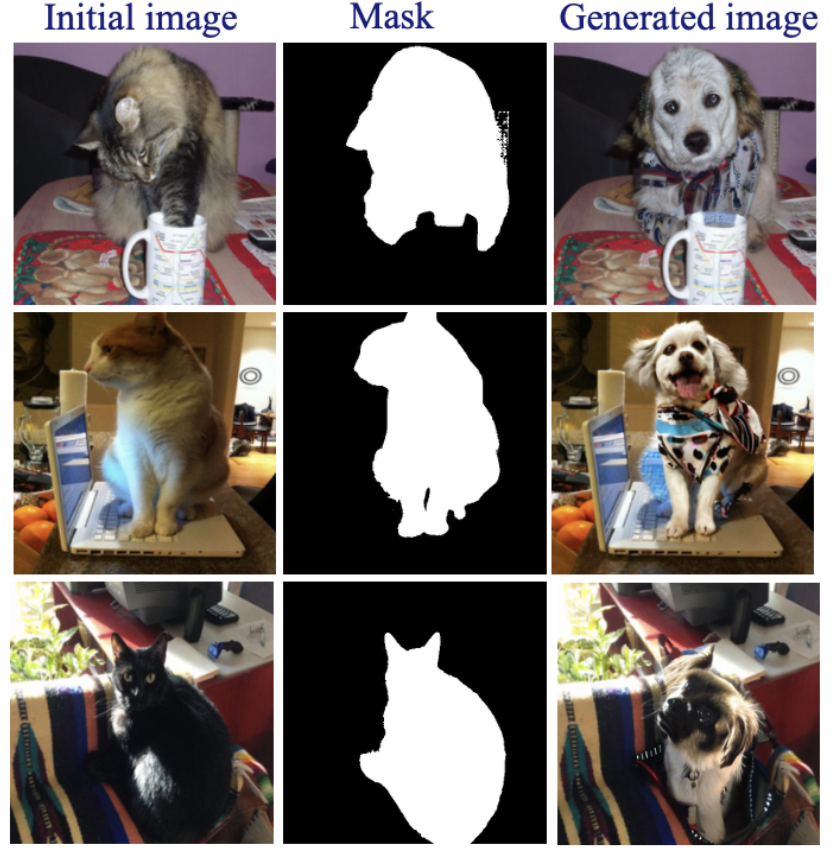}
    \caption{Examples of generated dog samples along with the corresponding masks and initial images. The initial images are belong to cat class.}
\label{fig:dognew}
\end{figure}

\begin{figure}[t]
    \centering
    \includegraphics[width=\linewidth]{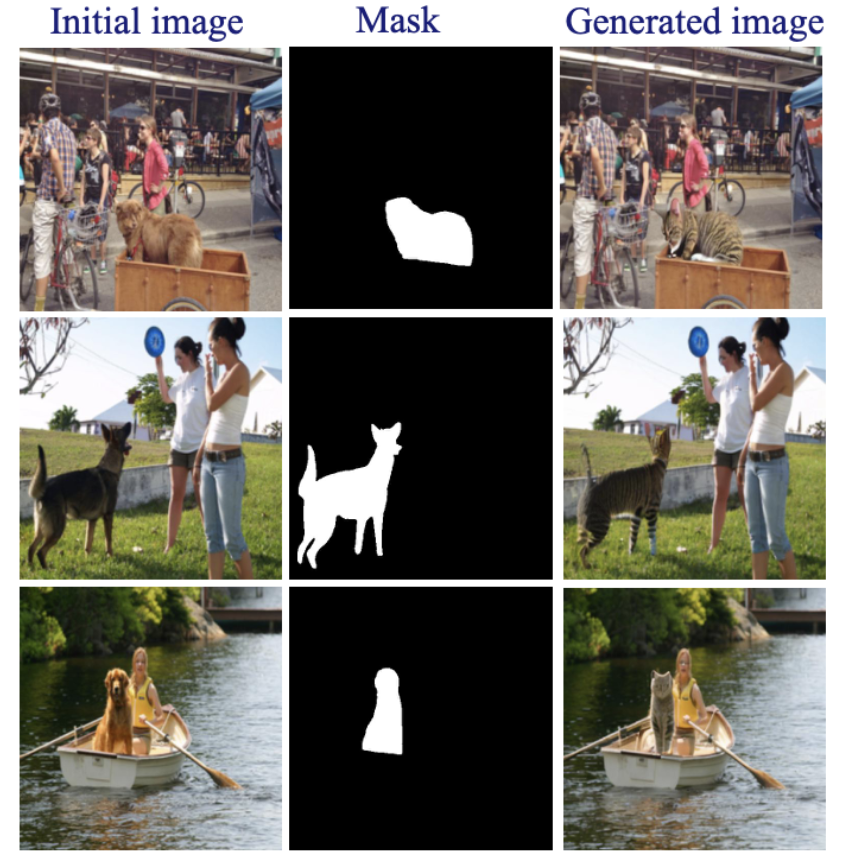}
    \caption{Examples of generated cat samples along with the corresponding masks and initial images. The initial images are belong to dog class.}
\label{fig:catnew}
\end{figure}

In this section, we provide additional qualitative examples across datasets. Figures~\ref{fig:landbirdnew} and~\ref{fig:waterbirdnew} present generated images from the Waterbirds dataset, Figures~\ref{fig:femalenew} and~\ref{fig:malenew} show results from the CelebA dataset, and Figures~\ref{fig:dognew} and~\ref{fig:catnew} show results from the Metashift dataset. Each figure showcases successful generations of minority group samples across different classes.

\end{document}